\documentclass{article}

    \PassOptionsToPackage{numbers, compress}{natbib}

\usepackage[preprint]{neurips22}

\usepackage[utf8]{inputenc} 
\usepackage[T1]{fontenc}    
\usepackage{hyperref}       
\usepackage{url}            
\usepackage{booktabs}       
\usepackage{amsfonts}       
\usepackage{nicefrac}       
\usepackage{microtype}      
\usepackage{xcolor}         
\usepackage{times}
\usepackage{latexsym}
\usepackage{amsmath}
\usepackage{graphicx}
\usepackage{multirow}
\usepackage{multicol}
\usepackage{float}
\usepackage{amssymb}
\usepackage{stmaryrd}
\usepackage{wrapfig}
\usepackage{subfig}
\usepackage{caption}
\usepackage{makecell}

\usepackage{amsmath}
\usepackage{subfig}
\usepackage{booktabs}
\usepackage{multirow}
\usepackage{amssymb}
\usepackage{xcolor}
\definecolor{darkgreen}{rgb}{0.0, 0.6, 0.2}

\newcommand{\TheName}{PSSL}

\title{Distilling Ensemble of Explanations for Weakly-Supervised Pre-Training of Image Segmentation Models}

\author{%
  Xuhong Li \\
  Baidu Inc. Beijing, China \\
  \texttt{lixuhong@baidu.com} \\
  \And Haoyi Xiong\thanks{Corresponding author} \\
  Baidu Inc. Beijing, China \\
  \texttt{xionghaoyi@baidu.com} \\
  \And Yi Liu \\
  Baidu Inc. Beijing, China \\
  \texttt{liuyi22@baidu.com} \\
  \And Dingfu Zhou \\
  Baidu Inc. Beijing, China \\
  \texttt{zhoudingfu@baidu.com} \\
  \And Zeyu Chen \\
  Baidu Inc. Beijing, China \\
  \texttt{chenzeyu01@baidu.com} \\
  \And Yaqing Wang \\
  Baidu Inc. Beijing, China \\
  \texttt{wangyaqing01@baidu.com} \\
  \And Dejing Dou \\
  Baidu Inc. Beijing, China \\
  \texttt{doudejing@baidu.com}
}

\begin{document}

\maketitle

\begin{abstract}

While fine-tuning pre-trained networks has become a popular way to train image segmentation models, such backbone networks for image segmentation are frequently pre-trained using image classification source datasets, e.g., ImageNet. Though image classification datasets could provide the backbone networks with rich visual features and discriminative ability, they are incapable of fully pre-training the target model (i.e., backbone+segmentation modules) in an end-to-end manner. The segmentation modules are left to random initialization in the fine-tuning process due to the lack of segmentation labels in classification datasets. In our work, we propose a method that leverages \emph{Pseudo Semantic Segmentation Labels} (PSSL), to enable the end-to-end pre-training for image segmentation models based on classification datasets. PSSL was inspired by the observation that the explanation results of classification models, obtained through explanation algorithms such as CAM, SmoothGrad and LIME, would be close to the pixel clusters of visual objects. Specifically, PSSL is obtained for each image by interpreting the classification results and aggregating an ensemble of explanations queried from multiple classifiers to lower the bias caused by single models. With PSSL for every image of ImageNet, the proposed method leverages a weighted segmentation learning procedure to pre-train the segmentation network \textit{en masse}. Experiment results show that, with ImageNet accompanied by PSSL as the source dataset, the proposed end-to-end pre-training strategy successfully boosts the performance of various segmentation models, i.e., PSPNet-ResNet50, DeepLabV3-ResNet50, and OCRNet-HRNetW18, on a number of segmentation tasks, such as CamVid, VOC-A, VOC-C, ADE20K, and CityScapes, with significant improvements. The source code is availabel at \url{https://github.com/PaddlePaddle/PaddleSeg}.

\end{abstract}

\section{Introduction}

Image semantic segmentation is a fundamental topic in computer vision, with broad applications in many domains, such as scene understanding, medical image analysis, robotic perception, etc.
In recent years, many approaches based on deep neural networks~\citep{zhao2017pyramid,chen2017rethinking,yuan2020object} have been developed and achieved remarkable performance on public datasets for image segmentation tasks, such as Pascal VOC~\citep{everingham2010pascal,hariharan2011semantic}, ADE20K~\citep{zhou2017scene} and Cityscapes~\citep{Cordts2016Cityscapes}.

\begin{figure*}[t]
\centering

\subfloat[Conventional Pre-Training on Classification for Image Segmentation]{\includegraphics[width=0.9\textwidth]{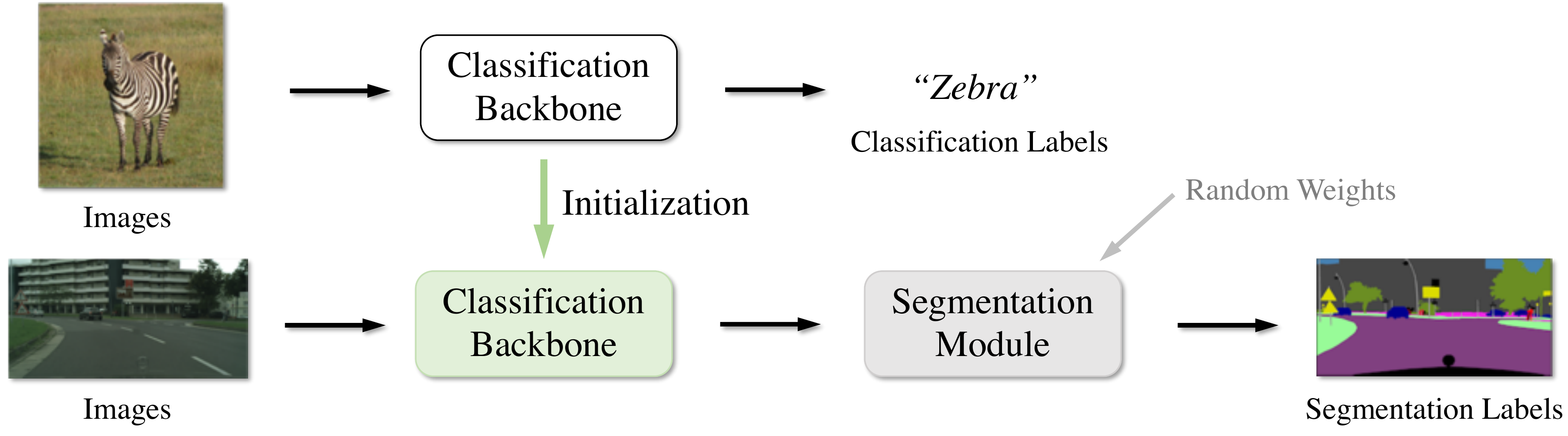}}\\
\subfloat[Pre-Training on PSSL for Image Segmentation]{\includegraphics[width=0.9\textwidth]{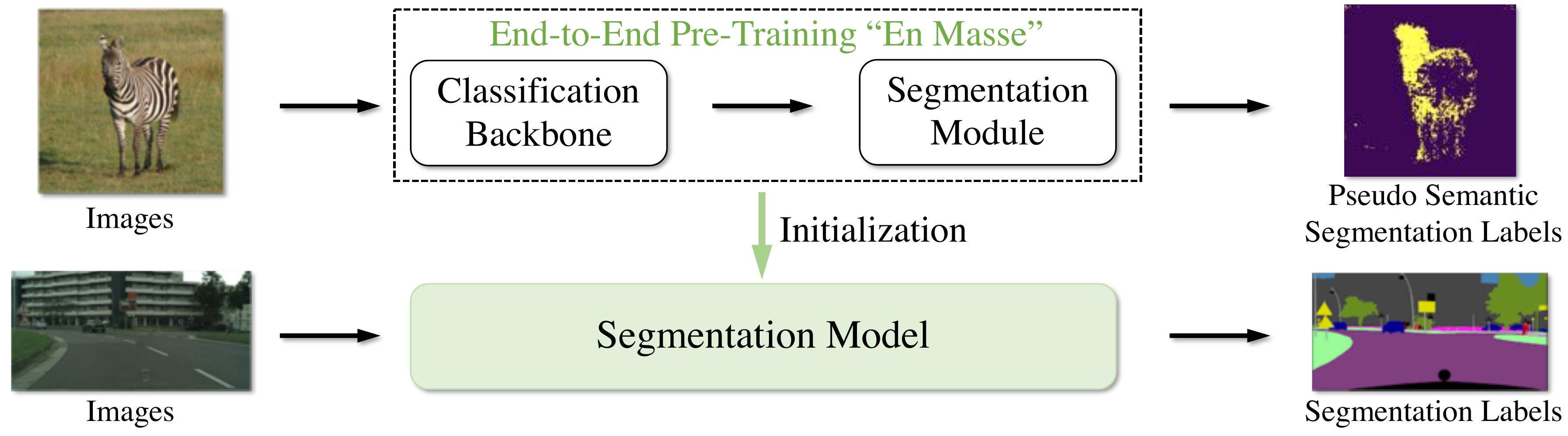}}

\caption{Illustration of the conventional (a) and our proposed (b) pre-training approaches. 
}
\label{fig:intro}
\end{figure*}

A conventional and practical training strategy for image segmentation, shown in Figure~\ref{fig:intro}(a), is to first adopt an image classification model as the backbone and subsequently incorporate with a segmentation module to make pixel-wise predictions\footnote{With more rigorous descriptions, the classification backbone is also modified to 
adapt to the segmentation task, such as the use of dilated convolutions~\citep{yu2016multiscale} and the decrease of convolution kernels' strides for reserving more spatial information.
}.
It then initializes the classification backbone with pre-trained weights from ImageNet~\citep{deng2009imagenet} or other large classification datasets but assigns random weights to the segmentation module as no prior given. 
Further, this strategy fine-tunes the whole model using datasets with segmentation labels to output pixel-wise predictions.

To close the gap between image classification and segmentation, some works~\citep{chen2017rethinking,zhao2017pyramid,zhang2020resnest} propose to fine-tune the whole segmentation model using the Microsoft COCO dataset with pixel-wise annotations~\citep{lin2014microsoft}.
This second round of pre-training is a practical approach to improving the performance on Pascal VOC because the COCO dataset covers all 20 categories of visual objects labeled in the Pascal VOC dataset.
These attempts show potential to further improve the image segmentation models with pre-training approaches, but it still needs a generalizable solution, especially when the source dataset does not cover the target domain.

\paragraph{Overview of Our Approach}
Our work is motivated to improve the imperfect \textit{status quo} of the fine-tuning process for image segmentation models, i.e., initializing the backbone using pre-trained weights of an image classification model, while leaving a large segmentation module with random initialization.
To this end, we follow the same settings of previous works~\citep{chen2017rethinking,zhao2017pyramid,zhang2020resnest,yuan2020object} that pre-train the image segmentation model using a classification dataset, e.g., ImageNet, where we propose to extract the pseudo semantic segmentation labels (PSSLs) from this classification dataset, and then use PSSLs to enable an end-to-end pre-training for image segmentation models.

Specifically, for every image in ImageNet, we \textbf{(i)} consider the explanation results of a deep model with respect to input features~\citep{ribeiro2016should,smilkov2017smoothgrad} as the highly important (super-)pixels potentially covering the visual objects, and \textbf{(ii)} aggregate the explanation results from multiple well-trained models and obtain PSSLs via the cross-model ensemble of explanations~\citep{consensus}, to reduce the bias induced by individual models.
With PSSLs, we propose to \textbf{(iii)} pre-train the backbone and the segmentation module \textit{en masse}, in an end-to-end manner, to enhance the image segmentation deep models, as illustrated in Figure~\ref{fig:intro}(b). 

The contributions of our work are summarized as follows:
\begin{itemize}
    \item We propose an end-to-end pre-training approach based on PSSLs, which is obtained through the ensemble of explanations, to pre-train a segmentation model using image classification datasets with image labels only. To the best of our knowledge, this pre-training strategy is the first end-to-end approach to pre-train segmentation models with classification datasets.
    \item We build the PSSL dataset, corresponding to over 1.2 million images in ImageNet using cross-model consensus without human involves. Both the PSSL dataset and the pre-trained segmentation models will be released for future researches and practical usages.
    \item Extensive experiments and thorough analyses have been conducted on five popular image segmentation tasks with three state-of-the-art models. Positive results with significant and consistent improvements confirm the effectiveness of the proposed approach.
\end{itemize}






\section{Related Work}

We review the related works from several directions and discuss our contributions compared to existing works.

\paragraph{Explanation Algorithms}
An important family of \textit{post-hoc} explanation algorithms~\citep{smilkov2017smoothgrad,ribeiro2016should,sundararajan2017axiomatic,bau2017network,selvaraju2017grad,ribeiro2018anchors} is to highlight the important input features that contribute the most to models' predictions.
In some specific scenarios, e.g., the classification in ImageNet~\citep{deng2009imagenet}, the important features are, in fact, the pixels of the visual objects, which align with image segmentation.
We do not focus on the explanation algorithms, but our proposed approach to generating the pseudo segmentation labels is based on the cross-model ensemble of \textit{post-hoc} explanations~\citep{consensus}.

\paragraph{Learning from Explanations}
Exploiting explanations for better training the deep model has been developed, mainly for two objectives, improving the model performance or providing better explanations.
To cite a few, Zagoruyko et al.~\cite{zagoruyko2016paying} imposed a regularizer to encourage a student network to learn the saliency maps from a teacher network for better knowledge distillation.
\cite{ross2018improving} proposed to train the model with regularizing the input gradients to be more interpretable explanations, for the objectives of improving the adversarial robustness and model interpretability.
\cite{chen2019looks} proposed to find prototypical patches for the final prediction and then train a model to align with these prototypes/explanations.
The learned model could achieve comparable accuracy and provide better explanations.
\cite{kim2020puzzle} proposed to improve Mixup~\cite{zhang2018mixup} by leveraging the explanation results.
Our approach is different from existing methods in two aspects: 
(1) We propose to use the cross-model ensemble of explanations~\cite{consensus}, which lowers the biases caused by individual models; 
(2) Rather than improving the original task, we transfer the explanations of classification models to pre-train models for segmentation. 

\paragraph{Deep Networks for ImageNet Semantic Segmentation}
As designing novel deep neural architectures has become a promising direction for specific learning tasks, most of the previous improvements on image semantic segmentation come from the network architecture expert designs (DeepLab~\cite{chen2017rethinking}, PSPNet~\cite{zhao2017pyramid}, OCRNet~\cite{yuan2020object} with different backbone networks).
Rather than designing new architectures, our approach proposes to leverage the hidden information from source data (mined by explanation algorithms), which is an orthogonal direction to architectural designs.
Any advances in network architecture could be complementary to the improvement made by our approach.
To confirm the advantage of our methods on top of various architectural designs, we conduct experiments on three different segmentation models and all of them obtain positive results, shown in Section~\ref{sec:exp}.

\paragraph{Weakly Supervised Semantic Segmentation (WSSS)}
WSSS aims at computing pixel-wise predictions with image-level annotations.
One relevant line of WSSS approaches~\citep{papandreou2015weakly,wei2018revisiting,lee2019ficklenet,zhang2020reliability} is to exploit explanation algorithms (e.g. CAM~\citep{bau2017network}, Grad-CAM~\citep{selvaraju2017grad}) to localize the objects in the image and to compute segmentation labels.
Our work does not predict pixel labels from coarse image annotations, but the approach to exploiting the explanation results is similar.
Another difference is that previous approaches generate the pixel-wise labels based on explanations of one single model, while we adopt the cross-model consensus of explanations across a number of deep models.
Advanced techniques from WSSS may be helpful to improve the accuracy of pixel-wise pseudo labels while we leave it as future work.

\paragraph{Self-Training \textit{vs} Pre-Training}
Self-training with unlabeled data~\citep{zoph2020rethinking} largely boosts the performance with enormous computation efforts.
It first trains a teacher model on human-labeled data, then generates soft labels on unlabeled data, and finally trains a student model jointly on human labels and soft labels.
However, this prohibitive self-training strategy solves specific tasks in an \textit{ad hoc} manner, meaning that the models are not shareable across different tasks.
Instead, our proposed approach is efficient and generalizable, which benefits from the pseudo pixel-wise labels from cross-model consensus of explanations and produces general segmentation models that enable performance boosts on various downstream tasks.

\paragraph{COCO Pre-Trained Models}
Microsoft COCO~\cite{lin2014microsoft} is a large dataset for image classification, object detection and image segmentation.
It contains 200K fine annotations across 80 categories, including the 20 objects in Pascal VOC. 
To benefit from this large dataset, many image segmentation algorithms~\cite{dai2015boxsup,zheng2015conditional,zhao2017pyramid,chen2017rethinking} improved the performance on Pascal VOC through selecting images from COCO that contain the 20 objects of Pascal VOC as extra training data.
This is a practical approach to boosting the performance on Pascal VOC, but it is less promising to apply on datasets that are not covered by the COCO dataset.
Our approach, however, proposes to pre-train the segmentation model on ImageNet with pseudo semantic segmentation labels, across 1000 different classes.
This enables a generalizable pre-training approach for various image segmentation tasks.

\paragraph{ImageNet Segmentation}
Guillaumin et al.~\cite{guillaumin2014imagenet} proposed to iteratively segment images, with bounding-box annotations and classification labels at the initial step.
Each step leverages the information from the previous step and refines the segmentation using GrabCut~\citep{rother2004grabcut}, a segmentation approach based on energy minimization.
They also conducted experiments to validate their method on ImageNet and released binary pixel-wise segmentation results of 4276 images across 445 classes.
Unlike using GrabCut, our approach, through cross-model ensemble of explanations, generates semantic segmentation labels for the entire training set of ImageNet.
The dataset will soon be released, containing over one million images covering all the 1000 classes in ImageNet, as detailed in the following section.
\section{Pseudo Semantic Segmentation Labels (\TheName{}) of ImageNet}

In this section, we introduce the procedure of generating \TheName{}
through the ensemble of explanations.

\begin{figure}[t]
\centering
\subfloat[][Cross-Model Ensemble of Explanations]{\includegraphics[width=0.8\columnwidth]{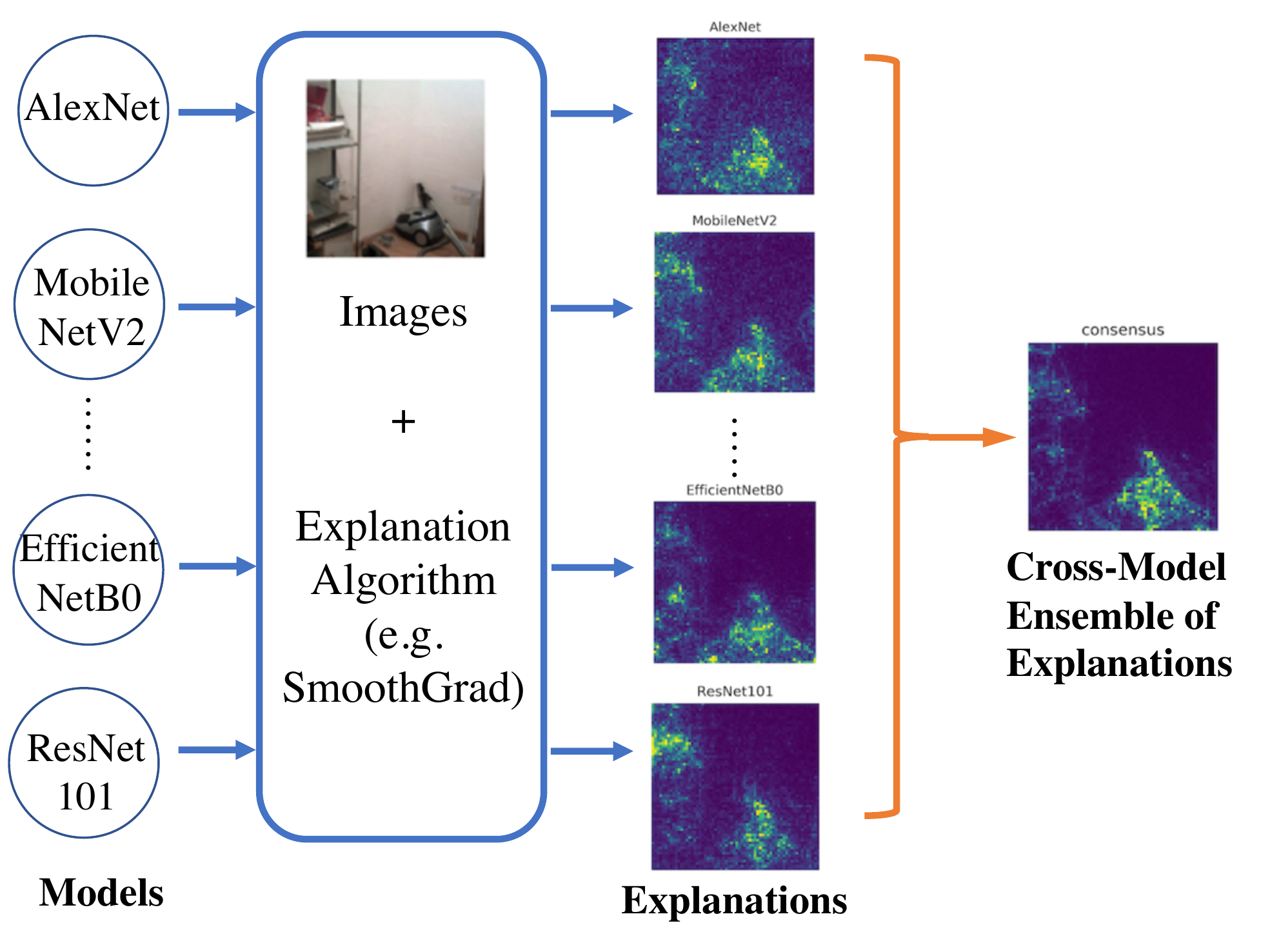}}\\

\subfloat[][Construction of PSSL]{\includegraphics[width=0.8\columnwidth]{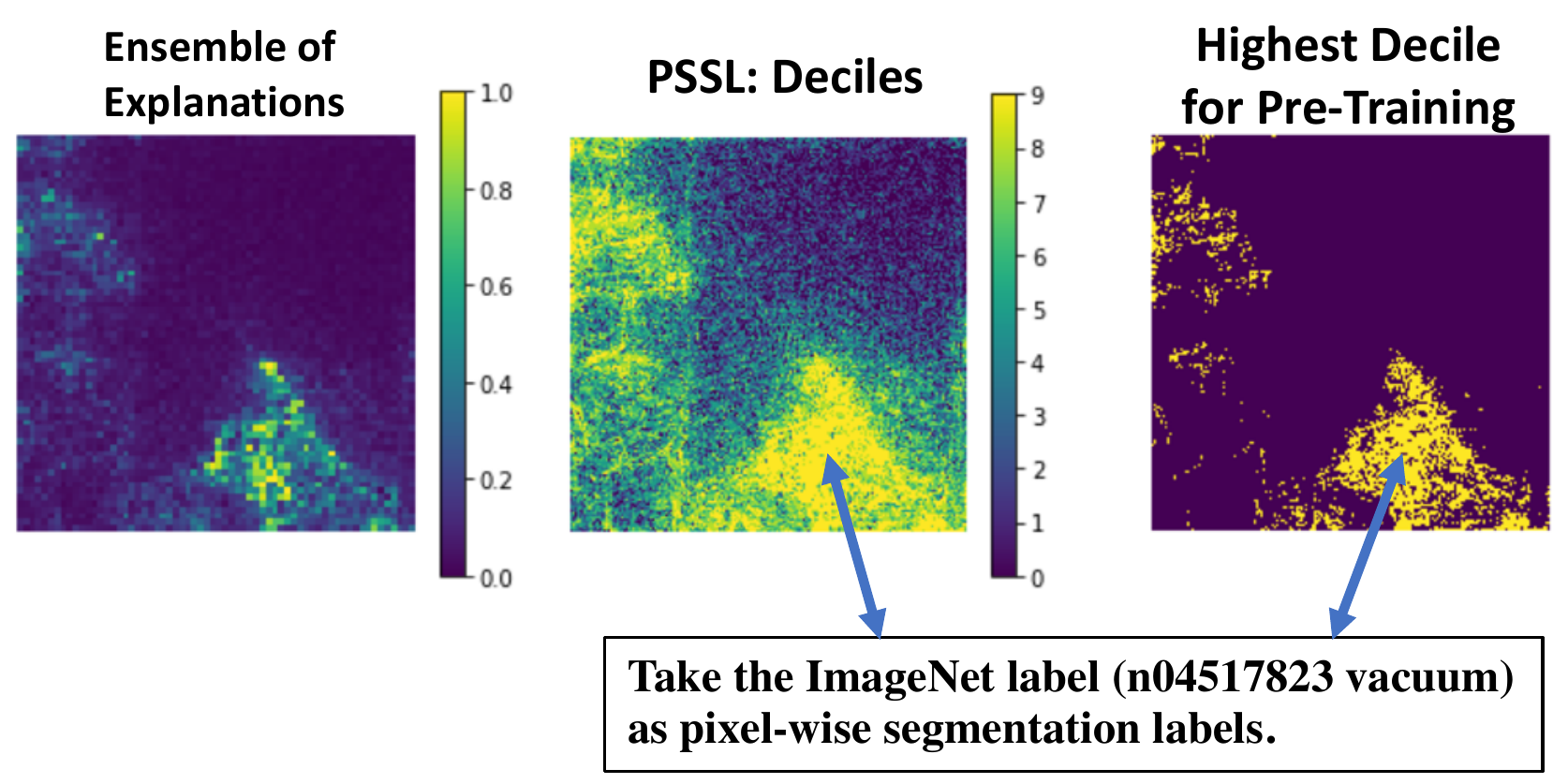}}
\caption{Illustration of (a) computing cross-model ensemble of explanations, (b) converting ensemble of explanations to images of deciles, and binarizing the highest decile.}
\label{fig:consensus}
\end{figure}

\subsection{Cross-Model Ensemble of Explanations}

We recall the approach of \textit{Consensus}~\citep{consensus}, i.e., cross-model ensemble of explanations.
As illustrated in Figure~\ref{fig:consensus}(a), \textit{Consensus} first collects a number of trained deep models, then adopts an explanation algorithm to interpret these deep models individually for each given image, and finally averages the explanation results across models.
As shown by~Li et al.~\cite{consensus}, the ensemble of explanations is well aligned to pixel-wise segmentation labels, and much better than single models.
In intuition, the averaged ensemble marginalizes out the variable of models, reducing biases from individual models.
Here, we generalize the idea of learning from explanations, and exploit the ensemble of explanations 
to enhance image segmentation pre-training.

\subsection{{Construction of PSSL}}

As introduced previously, the ensemble of explanations is much better aligned with image segmentation labels than individual ones.
Here, to produce pixel-level explanations for high resolution, we adopt the \textit{post-hoc} explanation algorithm SmoothGrad~\citep{smilkov2017smoothgrad}, while other appropriate algorithms may also be applicable here, such as Integrated Gradient~\citep{sundararajan2017axiomatic}, DeepLIFT~\citep{shrikumar2017learning} etc.
The number of models for the ensemble is suggested to be larger than 15 to get a good alignment with segmentation~\citep{consensus}.
So the first step to constructing PSSL is based on the ensemble of SmoothGrad explanations across 15 deep models, to obtain the contribution scores of pixels for every image in the training set of ImageNet,
\begin{equation}
    S(x) = 1/M \sum_{i=1}^{M} S_i(x),
\end{equation}
where $S_i(x)$ is the score of pixel $x$ given by the explanation algorithm w.r.t. the $i$-th model, rescaled by a min-max normalization, and $M$ is the total number of models.

PSSLs with floating point take much space in disk.
So the second step of the construction, with the objective of reducing the file size without losing much information, is to perform a quantization process to convert quasi-continuous scores (floating point 32 bits) to deciles (0-9).
Specifically the decile rank $D$ of pixel $x$ is computed by 
\begin{equation}
    \label{eq:decile}
    D(x) = \lfloor Z(x) / N * 10 \rfloor,
\end{equation}
where $ Z(x) = \lvert \{ y\ \vert\ S(y) < S(x) \} \rvert $, is the number of pixels that have lower scores than $x$, $N$ is the total number of pixels and $ \lfloor F \rfloor $ is the largest integer that is less than $F$.
We denote the set of $\{x \vert D(x) = i \}$ as $D_i$, where $i=0,1,2,...,9$.
In fact, practically we take the highest decile $D_9$ as the segmentation labels for pre-training.

The last step is to connect the deciles to category labels.
Image-wise category labels can be directly loaded from ImageNet, and by the reason that almost every image in ImageNet contains only one main object, pseudo pixel-wise labels can be reasonably assigned by the image-wise label.
In our setting, the labels $T(D_9) = T(\mathcal{I})$, where $T(\mathcal{I})$ is the image-wise category from ImageNet of the image $\mathcal{I}$.

The three steps are illustrated in Figure~\ref{fig:consensus}(b).
In summary, PSSL is created by repeating these steps to obtain over one million images of deciles corresponding to images in the training set of ImageNet.
Currently PSSL does not contain a validation/test set, because the pseudo labels have not been manually verified or corrected.
The evaluations on them do not make much sense.
We thus suggest using the dataset of PSSL for designing the pre-training strategies only, since the effectiveness of PSSL-pre-trained models is validated on downstream tasks, with experiments presented in Section~\ref{sec:exp}.
Further usages of the pre-trained model are planned as future work.

\section{Proposed Approach: End-to-End Pre-Training on PSSL} 
\label{sec:approach}

In this section, we introduce the proposed end-to-end pre-training approach for segmentation models based on the dataset of PSSL.

\subsection{Take the Cream and Dross}
Pseudo labels from PSSL are not guaranteed to be accurate.
Precisely filtering out the noises is essential to generally improve the dataset's quality and effectively enhance the segmentation models with PSSL.
Here we consider two directions to cope with the noises.

For practical efficiency, we choose to take a constant decile for all pseudo labels, i.e., the highest one of pixels $D_9$, as segmentation (pseudo) ground truth for pre-training.
This threshold works well in practice, compared to using more deciles.
This may be explained by that using more deciles would introduce more noises to the supervision for the images in which the sizes of objects are small.
More advanced techniques will surely improve the effectiveness.
This is also the reason that we release the PSSL as images of deciles instead of binarized images, for future researches on adaptive approaches to choosing the threshold for individual images, or on algorithms of refining the segmentation labels based on other prior information.

The cream is taken; the dross, however, cannot be directly discarded.
Simply dropping the pixels of $D_{i<9}$ during training would boil down the trained segmentation model to a trivial classification model, predicting all pixels in one image to the same category.
Training with the background class essentially helps to segment images semantically.
Meanwhile, we also tried to ignore $D_8$, leaving $T(D_9) = T(\mathcal{I})$ and the rest as background, which decreased the performance on downstream tasks.


\subsection{Imbalanced Learning Problem}

As previously introduced, to reduce noises, for each image $\mathcal{I}$, we set $T(D_9) = T(\mathcal{I})$ and $T(D_{i<9}) = B$, where $B$ is the index of the background class.
Considering the large ratio of background pixels, this training problem is essentially an imbalanced learning problem of classifying pixels.
Pixels of background are around 9 000 times as many as pixels of any other category, in the case of 1 000 balanced categories.
Several advanced loss designs for the imbalanced classification are considered here, such as Focal Loss~\citep{lin2017focal} and Loss Max-Pooling~\citep{bulo2017loss}.
However, the imbalance in PSSLs also involves the noises.
Direct applications of Focal Loss or others may give high weights to noisy pixels, leading to undesired effects.
So, instead of adopting such advanced technologies, we consider the weighted cross entropy objective function:
\begin{equation}
    l(x) = \sum_{c=1}^{K} -\ w_{c}\ t_{c,x}\ \log (p_{c,x}),
\end{equation}
where $K$ is the number of classes, $w_{c}$ is the weight for class $c$, $t_{c,x}$ is the one-hot label at $c$ converted from $T(x)$, $p_{c,x}$ is the predicted probability of class $c$ for pixel $x$.

Through preliminary experiments, we have found that set $w_B$ to $0.1$ and others $w_{i\neq B}$ to 1 works well in practice.
We take parts of \TheName{} (50K and 500K pseudo-labeled samples respectively), train a DeepLabV3-ResNet50 segmentation model and fine-tune the trained segmentation model on the PASCAL VOC segmentation dataset~\cite{everingham2010pascal,hariharan2011semantic}.
The experiment setups here are the same as those in Section~\ref{sec:exp}, except that \textit{the training epoch is less here, leading to slightly lower mIoU scores compared to the main results}.
We vary the background weight $w_B$ without changing others to show the effects of different background weights on Pascal VOC.
The results in Table~\ref{tab:bg-weight} indicate that $0.1$ is the best among the grids.

\begin{table}[h]
\caption{Performance with respect to the background weight in the objective function during pre-training with \TheName{}-50K and \TheName{}-500K, evaluated on the PASCAL VOC 2012 validation set using DeepLabV3-ResNet50.}
\label{tab:bg-weight}
\centering
\begin{tabular}{@{}lllll@{}}
\toprule
\textbf{BG Weight} & 0.001 & 0.01  & 0.1  & 1.0   \\ \midrule
\TheName{}-50K       & 72.7 & 73.9 & 74.1 & 73.7 \\
\TheName{}-500K       & 75.4 & 76.1 & 76.7 & 74.3 \\ \bottomrule
\end{tabular}
\end{table}
\subsection{Data Matter}

Training with more correctly labeled samples is more probable to produce a good model. With pseudo-labeled image segmentation samples, this still holds.
We use separately 50K, 200K, 500K, and 1M pseudo segmentation labels from PSSL to pre-train the segmentation models and obtain four pre-trained models.
Then we fine-tune them on Pascal VOC and compare them with the conventional fine-tuning approach. The results in Table~\ref{tab:sg-num} clearly show the trend of increase with more samples used during pre-training. These results also indicate that with 500K pseudo labels, the performance on the downstream task is slightly better than the conventional fine-tuning approach, which only initializes the backbone part of the segmentation model.

\begin{table}[h]
\caption{Performance with respect to the sample number of PSSL, evaluated on the PASCAL VOC 2012 val set using DeepLabV3-ResNet50. ImageNet: The backbone is pre-trained with the ImageNet classification labels. \TheName{}: The backbone is pre-trained on different numbers of PSSL samples generated by the cross-model ensemble of explanations. 
}
\centering
\begin{tabular}{@{}l|c|llll@{}}
\toprule
\begin{tabular}[c]{@{}l@{}}Pre-Training\\ Dataset\end{tabular} & \begin{tabular}[c]{@{}l@{}}Image-\\ Net\end{tabular}  & \multicolumn{4}{c}{PSSL}                                                                                                                                                                                                                                                                             \\ \midrule
Nb. Samples                                             & 1.2M                                                   & 50K                                                                  & 200K                                                                 & 500K                                                                       & 1M                                                                         \\ \midrule
mIoU                                                    & 76.2                                                 & \begin{tabular}[c]{@{}l@{}}74.1 \\ ({\color{red} -2.1})\end{tabular} & \begin{tabular}[c]{@{}l@{}}75.4 \\ ({\color{red} -0.8})\end{tabular} & \begin{tabular}[c]{@{}l@{}}76.7 \\ ({\color{darkgreen} +0.5})\end{tabular} & \begin{tabular}[c]{@{}l@{}}77.8 \\ ({\color{darkgreen} +1.6})\end{tabular} \\ \bottomrule
\end{tabular}
\label{tab:sg-num}
\end{table}



\section{Experiments}
\label{sec:exp}

 \begin{table}[t]
\centering
\caption{Comparison on five segmentation datasets with three popular segmentation models measured in mIoU, where ImageNet indicates the conventional initialization of using ImageNet-pretrained classification model for segmentation, and \TheName{} indicates that the weights are initialized from the pre-trained model on the dataset of PSSL. In green are the gaps of at least {\color{darkgreen}+\textbf{0.6}} points.}

\resizebox{\columnwidth}{!}{
\begin{tabular}{@{}l|c|c|c|c|c|c@{}}
\toprule
           & \multicolumn{2}{c|}{PSPNet-ResNet50} & \multicolumn{2}{c|}{DeepLabV3-ResNeSt50} & \multicolumn{2}{c}{OCRNet-HRNetW18} \\ \cmidrule(l){2-7} 
           & ImageNet  & \TheName{}                  & ImageNet          & \TheName{}             & ImageNet        & \TheName{}      \\ \midrule
CamVid     & 65.9      & 68.1 {\color{darkgreen}\footnotesize(+\textbf{2.2})}  & 66.6              & 69.1 {\color{darkgreen}\footnotesize(+\textbf{2.5})}          &   59.2              &    62.8 {\color{darkgreen}\footnotesize(+\textbf{3.6})}          \\
VOC-A        & 79.4      & 80.3 {\color{darkgreen}\footnotesize(+\textbf{0.9})}  & 79.1              & 80.1 {\color{darkgreen}\footnotesize(+\textbf{1.0})}          &  76.4               &  77.1 {\color{darkgreen}\footnotesize(+\textbf{0.7})}             \\
VOC-C      & 47.0      & 48.5 {\color{darkgreen}\footnotesize(+\textbf{1.5})}  & 48.8              & 49.4 {\color{darkgreen}\footnotesize(+\textbf{0.6})}          &  44.5               &  45.7 {\color{darkgreen}\footnotesize(+\textbf{1.2})}             \\
ADE20K     & 42.9      & 43.8 {\color{darkgreen}\footnotesize(+\textbf{0.9})}  & 45.2              & 45.8 {\color{darkgreen}\footnotesize(+\textbf{0.6})}          &  40.0               &  40.9 {\color{darkgreen}\footnotesize(+\textbf{0.9})}             \\
Cityscapes & 78.7      & 78.9 {\footnotesize(+0.2)}  & 79.0              & 79.7 {\color{darkgreen}\footnotesize(+\textbf{0.7})}          &     79.6            &       79.8  {\footnotesize(+0.2)}        \\ \bottomrule
\end{tabular}
}
\label{tab:main-results}
\end{table}

To address the imperfect parameter initialization for image segmentation, we propose to pre-train the segmentation models en masse on PSSL.
These pre-trained models are then used as initial weights for the evaluations on five downstream segmentation tasks through fine-tuning.
We compare with the conventional approach of initializing the classification backbone only, so as to validate the effectiveness of our proposed end-to-end pre-training strategy.
This section, therefore, presents the experiments of pre-training on PSSL and fine-tuning on downstream tasks.

\subsection{Models}
Our proposed pre-training strategy is independent of network structures.
To experimentally demonstrate this independence, we conduct experiments on three models that are different in both backbone models and segmentation modules.
Specifically, we consider three popular segmentation models: PSPNet~\cite{zhao2017pyramid}, DeepLabV3~\cite{chen2017rethinking} and OCRNet~\cite{yuan2020object}, with ResNet50~\cite{he2016deep}, ResNeSt50~\cite{zhang2020resnest} and HRNetW18~\cite{wang2020deep} as backbones respectively.
These three models improve the image segmentation based on different expert designs on network structure, while our approach further boosts the performance by addressing the initialization issue of segmentation models in an orthogonal way.


\subsection{Pre-Training Experiments}

Following the approach introduced in Section~\ref{sec:approach}, we conduct the pre-training experiments with the three models.

Some experiment details are added here. 
The input image size to the network is 256, a similar value as used in image classification, for getting a large batch size 64\footnote{Due to the limited GPU memory, a larger input size for training like 420 would reduce the batch size to 16. While our used setting produces similar results to (marginally better than) this setting, our setting needs less wall time for the same number of epochs.}.
The training epoch is 30.
The learning rate is 0.01 and decreased to 0.001 at 20$^{th}$ epoch.
The data augmentation methods for general image segmentation are also used here, such as random scales, crops, flipping, blurs, rotations etc.
Contrary to standard training processes, the model is not desired to be well converged, avoiding overfitting to the noises in PSSL.
So an early stop at 30$^{th}$ epoch is performed.
The source code is also available for unmentioned details and reproduction purposes.

This setting costs 3 to 4 days on an 8-V100 server for each of the three models.
Compared to the conventional fine-tuning (with using public available pre-trained classification models as initialization), this is the only additional computation required for adopting our proposed method.
We also remark that with the released pre-trained segmentation models, even such additional computations are not required for future research and practical usages.

\subsection{Downstream Segmentation Datasets}
We evaluate the pre-trained segmentation models on five datasets, i.e., CamVid~\citep{brostow2009semantic}, Pascal VOC Augmented (shortly VOC-A)~\citep{everingham2010pascal,hariharan2011semantic}, Pascal VOC Context (VOC-C)~\citep{everingham2010pascal}, ADE20K~\citep{zhou2017scene} and Cityscapes~\cite{Cordts2016Cityscapes}.
Each of them is attached with an image segmentation task, in different scenarios.
In general, CamVid is a small dataset with 367 images for training, while others have more than thousands of training images.
Two of them are based on driving scenes, i.e., CamVid and Cityscapes;
Two of them focus on general object segmentation, i.e. two VOC datasets;
ADE20K involves various scenes and contains 150 objects, many more than the number of classes in other datasets;
Cityscapes provides high-resolution images of 2048$\times$1024, while other datasets contain images with the larger edge being around 500.
Detailed dataset information, including sizes of the training/validation/test sets, resolution, number of classes to classify, can be found in the appendix.

The idea is to evaluate the PSSL-pretrained models on datasets of different scales, scenes and objects.
Consistent improvements are observed across datasets.

\subsection{Implementation Details}
We have three models on five datasets to evaluate.
Generally, the best hyper-parameters are not the same across the 15 dataset-model combinations.
The general rule to choosing the hyper-parameters for fair comparisons is that in each combination, the hyper-parameters are the same, or the way to tuning them is the same.
The only hyper-parameter to tune in our experiments is the initial learning rate.
The strategy for the best initial learning rate is to search the peak value as the best one from \{$5,2,1 \times 10^{-n}, \text{where } n = 1,2,3,...  $ \}\footnote{Specifically, the values of \{1e-4, 2e-4, 5e-4, 1e-3, 2e-3, 5e-3, 1e-2, 2e-2, 5e-2\}.} have been searched.
We show the tuning process on VOC-A in the appendix.
Other hyper-parameters are fixed within each combination, such as batch size (16), weight decay (1e-3 for CamVid, 1e-4 for others), polynomial-decay learning rate policy, base size (1024 for Cityscapes, 520 for others), crop size (864 for Cityscapes, 480 for others), epochs (30 for CamVid, 50 for both VOC datasets, 180 for ADE20K-DeepLabV3, 100 for the rest), with all commonly-used data augmentation methods, and so on.
Multi-scale evaluations are performed.
More details can be found in the source code.

\subsection{Main Results}

The main results of three models on five datasets are shown in Table~\ref{tab:main-results}, where the proposed PSSL-pretraining strategy outperforms the conventional fine-tuning method with ImageNet-classification models on all dataset-model combinations.
The mIoU scores are increased by at least 0.9 points for 9 out of 15 combinations, knowing that we did not use additional data in both source and target domains.

We observe huge gains on CamVid across the three models.
This is because prior information is very beneficial for tasks with few training data points (there are only 367 training images on CamVid), and this also reveals that our proposed pre-training strategy effectively initializes the segmentation models.

Moreover, our method shows significant improvements for relatively large datasets.
The two VOC datasets and ADE20K generally contain similar images to ImageNet, in sens of both image resolution and the targeted visual objects.
An increase of around one point is observed for most of them across different models on these three datasets.

Our method also provides positive yet variant results on Cityscapes.
Note that the PSSL-pretrained models take images of 256 as input, while the input image size is 864 for training on Cityscapes.
Image resolutions are also quite different in ImageNet and Cityscapes.
Nevertheless, the PSSL pre-training strategy is still effective in such scenarios, while yielding marginal improvements for PSPNet and OCRNet.
This may indicate that the image resolution is an essential factor that affects the transfer from source to target domains. 
This is also an issue in the conventional fine-tuning approach, and we leave this as future work.

\section{Analysis on PSSL-Pretrained Segmentation Models}

The segmentation models pre-trained on PSSL have been proved to be good initial points for fine-tuning on various downstream datasets.
In this section, we provide analyses of these pre-trained models and show their potentials.

\begin{figure}[t]
\centering
\includegraphics[width=\columnwidth]{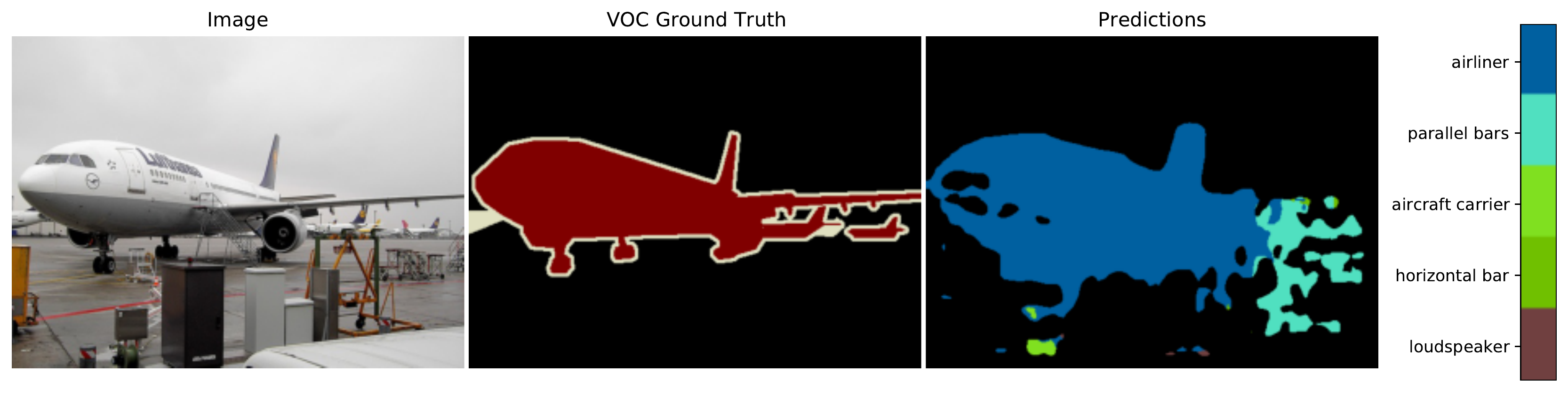}
\includegraphics[width=\columnwidth]{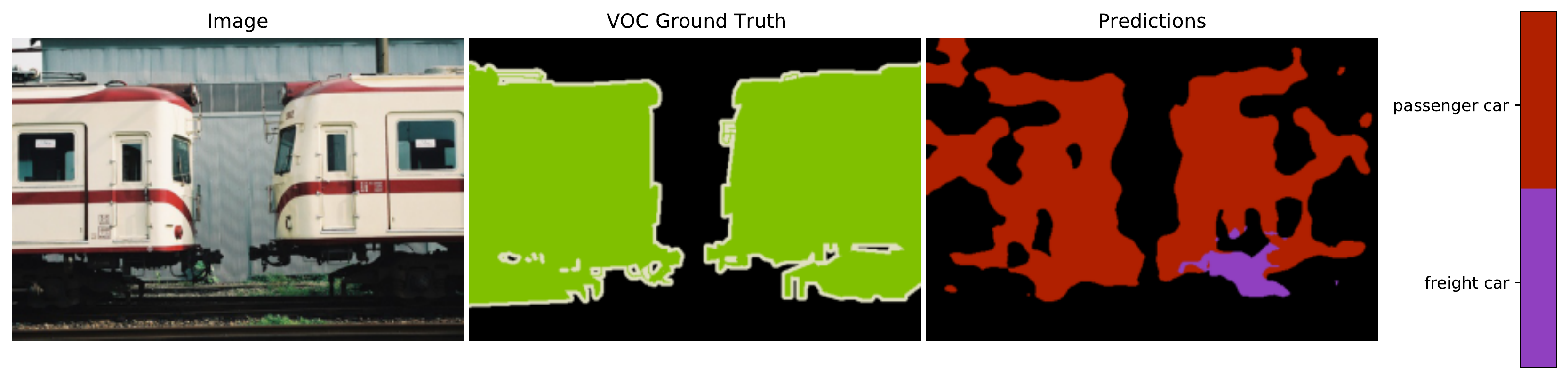}
\includegraphics[width=\columnwidth]{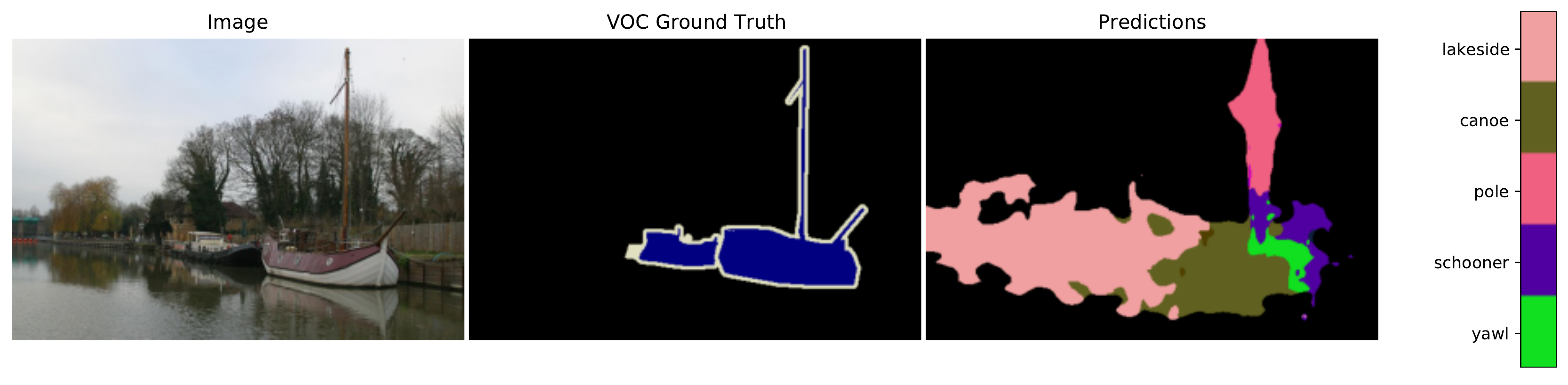}
\includegraphics[width=\columnwidth]{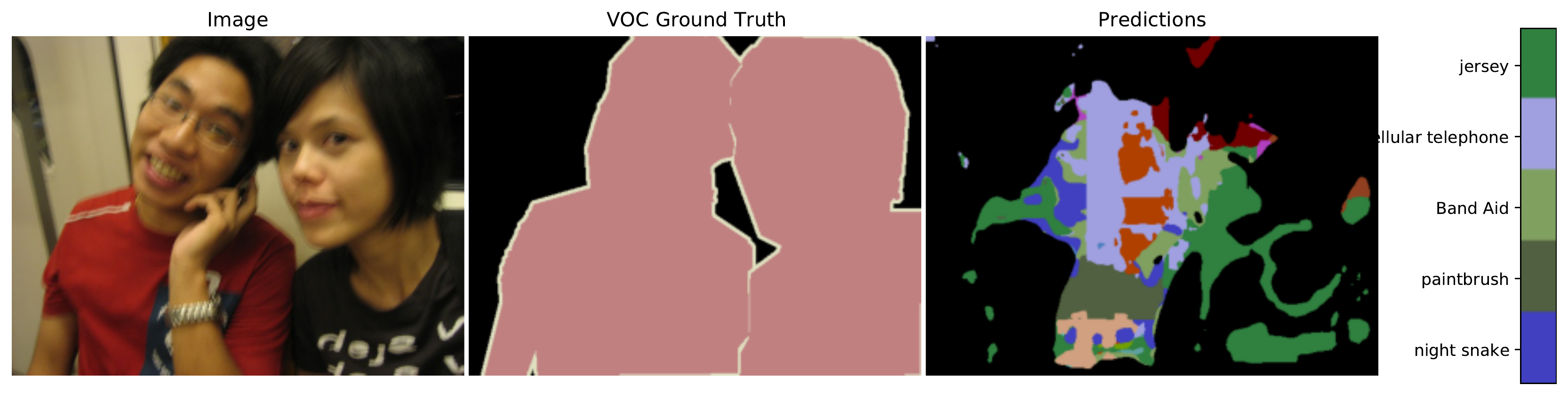}

\caption{Predictions from PSSL-pretrained segmentation models. The four columns are respectively original images, pixel-wise ground truth from VOC-A, prediction results from PSSL-pretrained models, and legends for the top categories. Best viewed in color and with zoom in.}
\label{fig:vis}
\end{figure}

\subsection{Classifying Images}

We first evaluate the pre-trained segmentation models on image classification tasks, by simply taking the prediction probability averaged from all pixels as the final prediction for image classifications.
Single-scale evaluation experiments are performed on the ImageNet validation set, which was not used for training.
The results of top-1 accuracy are reported in the second row of Table~\ref{tab:converge}, compared to the ImageNet-classification pre-trained model that gets 78.7\%.

In the main experiments, an early stopping at 30$^{th}$ epoch was adopted for pre-training on PSSL, to avoid converging to the noises of pseudo pixel-wise labels.
As a result, the convergence to the image labels is not reached.
That may be the reason for the decrease in classification accuracy.
As a remedy, we continue to train the segmentation model for another 30 epochs on PSSL.
That gives a comparable classification accuracy on ImageNet, but this well converged segmentation model produces slightly lower mIoU scores on VOC-A than the early-stopped one, as reported in the last row of Table~\ref{tab:converge}.
Segmentation experiments follow the same configuration as in the previous section.

\begin{table}[h]
\caption{Comparison of PSSL-pretrained PSPNet-ResNet50 with 30 and 60 epochs and ImageNet-pretrained ResNet50, evaluated using the pre-trained models on ImageNet classification and the fine-tuned models on VOC-A segmentation.
}
\label{tab:converge}
\centering
\begin{tabular}{@{}lrr|r@{}}
\toprule
\begin{tabular}[c]{@{}l@{}}Epochs of PSSL-\\ Pretraining \end{tabular} & 30   & 60   & \begin{tabular}[c]{@{}r@{}}ImageNet-\\ Pretrained\end{tabular} \\ \midrule
ImageNet (Top-1 Acc)                                                  & 77.6 & 78.6 & 78.7                                                            \\
VOC-A (mIoU)                                                          & 80.3 & 80.0 & 79.4                                                            \\ \bottomrule
\end{tabular}
\end{table}

Generally, better (ImageNet-)pretrained models transfer better~\citep{kornblith2019better}.
However, in our case, the source dataset of PSSL contains some amount of noise.
Better PSSL-pretrained models do not necessarily learn richer representations; instead, they may have larger potentials of remembering these noises, which is not profitable for fine-tuning.
For instance, a decrease of 0.3 points in mIoU scores is observed in Table~\ref{tab:converge}.
To preclude this and be more efficient, we used the early-stopped models in the main experiments.

\subsection{Segmenting Images to 1000 Categories}
The pre-trained segmentation models are capable of classifying pixels into 1000 categories plus an additional background class.
We visually show the potentials of directly using these models to perform the segmentation task on VOC-A.
Figure~\ref{fig:vis} shows the prediction results of PSPNet-ResNeSt50, for the first (three) images from the VOC-A validation set, plus the first image that contains ``person''.
Due to the absence of ``person'' in ImageNet, the pre-trained model is not able to correctly segment the pixels.
Instead, the model recognizes the clothes and the telephone.
Short-version category names are loaded\footnote{For complete label names, refer to the list on GitHub: \url{https://gist.github.com/yrevar/942d3a0ac09ec9e5eb3a}.}.
More examples can be found in the supplementary materials.

\section{Conclusion and Future Work}
We proposed an end-to-end pre-training strategy for complex scene understanding and parsing problems.
Conventional fine-tuning for the segmentation model takes a pre-trained classification model as the backbone but leaves the segmentation module to random initializations.
To address this issue, we created a new dataset of pseudo semantic segmentation labels, named PSSL, containing over one million pseudo labels, through the cross-model ensemble of explanations.
We pre-trained three popular segmentation models on PSSL with ImageNet, and used them as initial weights for fine-tuning on five downstream segmentation tasks.
Experiments showed positive results with significant improvements, demonstrating the effectiveness of the proposed end-to-end pre-training strategy.
We furthermore provided analyses on the pre-trained models, indicating the potentials and other possible usages of PSSL-pretrained models.

Future work includes but is not limited to WSSS, better segmentation with more robust priors, self-training with unlabeled data using an ensemble of explanations, etc.
We hope the dataset and the pre-trained models (publicly available soon) can be helpful for the related domains.

\bibliographystyle{unsrtnat}
\bibliography{lxh.bib}


\clearpage


\appendix

\begin{table*}[h]
\centering
\caption{Dataset information.}
\begin{tabular}{@{}lrrrrr@{}}
\toprule
           & Tr. Set & Val Set & Resolution       & Nb. Clas.  & Info.          \\ \midrule
CamVid     & 367          & 233     & 480$\times$360   & 11 (w/o bg)  & Driving Scenes \\
VOC-A      & 10582        & 1449    & $\le$ 500        & 21 (w/ bg)   & Common Objects \\
VOC-C      & 4998         & 5105    & $\le$ 500        & 60 (w/o bg)  & Common Objects \\
ADE20K     & 20210        & 2000    & $\approx$ 500           & 150 (w/o bg) & Various Scenes \\
Cityscapes & 5000         & 500     & 2048$\times$1024 & 19 (w/o bg)  & Driving Scenes \\ \bottomrule
\end{tabular}
\label{tab:datasets}
\end{table*}

\section{Dataset Information}

We evaluate three popular segmentation models on five datasets, i.e., CamVid~\citep{brostow2009semantic}, Pascal VOC Augmented (shortly VOC-A)~\citep{everingham2010pascal,hariharan2011semantic}, Pascal VOC Context (VOC-C)~\citep{everingham2010pascal}, ADE20K~\citep{zhou2017scene} and Cityscapes~\cite{Cordts2016Cityscapes}.
These datasets are of various scales, scenarios, and numbers of labeled objects.
Table~\ref{tab:datasets} shows the details of these five datasets.
The idea is to evaluate the PSSL-pretrained models on various datasets.
Experiments show consistent improvements across these datasets.

\section{Learning Rate Tuning}
\begin{figure}[t]
\centering
\includegraphics[width=0.45\columnwidth]{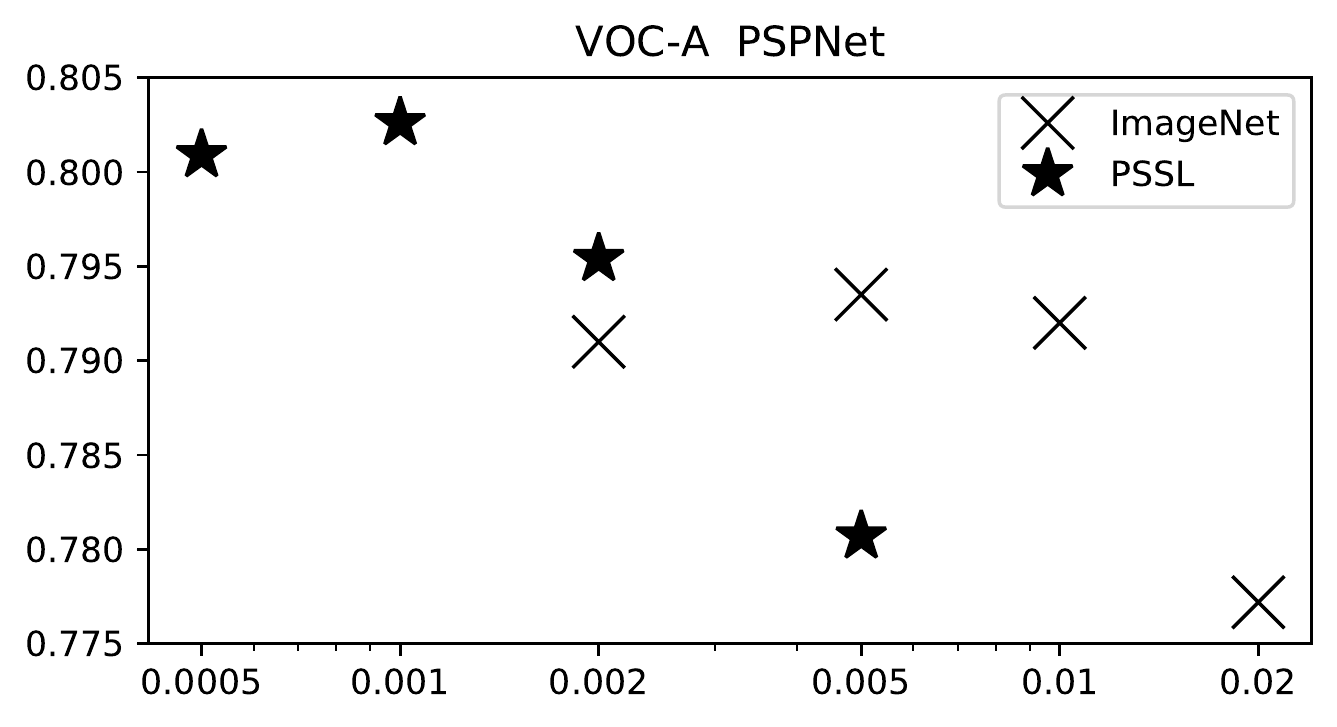}
\includegraphics[width=0.45\columnwidth]{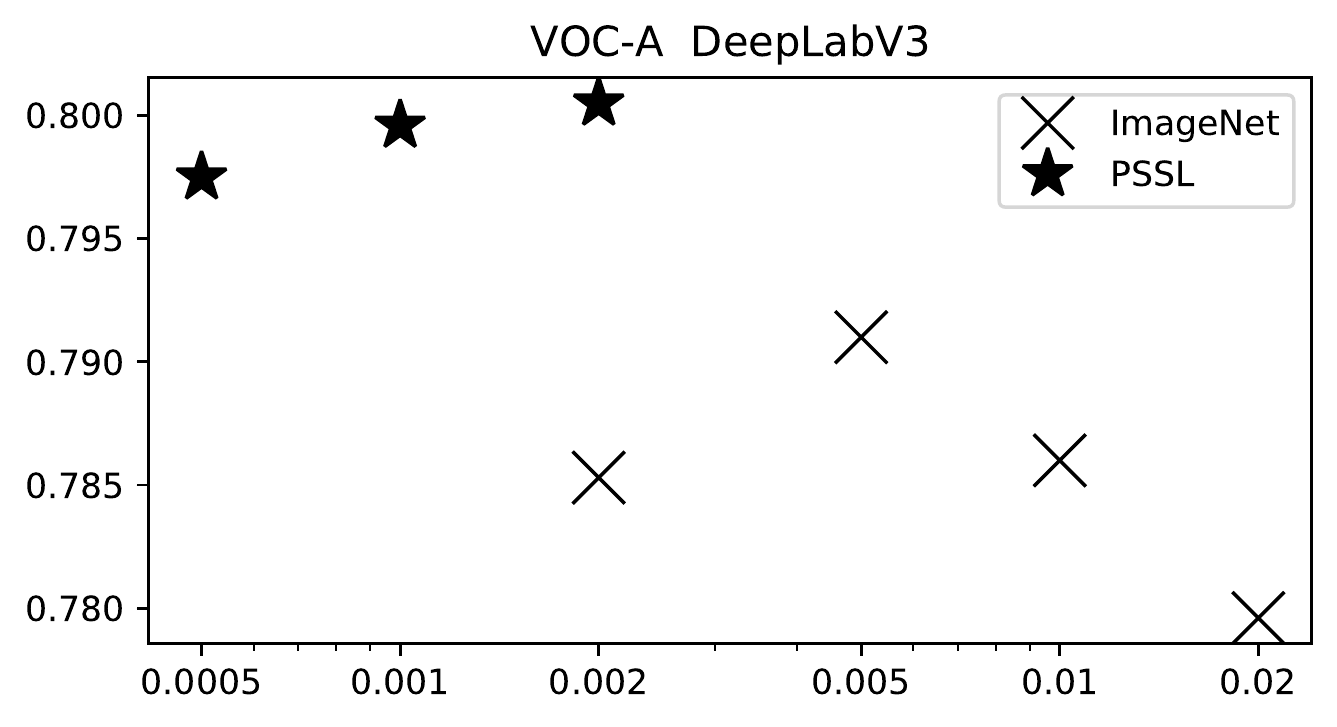}

\caption{Learning rate tuning results on VOC-A. Similar results can be found on other dataset-model combinations.}
\label{fig:tuning}
\end{figure}

We recall the hyper-parameter tuning rule in our experiments.
The only hyper-parameter to tune in our experiments is the initial learning rate, and the strategy for the best initial learning rate is to search the peak value as the best one from \{$5,2,1 \times 10^{-n}, \text{where } n = 1,2,3,...  $ \}.
To visualize the tuning steps of our experiments, we present the results on VOC-A~\citep{everingham2010pascal,hariharan2011semantic}.
As shown in Figure~\ref{fig:tuning}, we stopped tuning when we found a clear peak for each setting, for both ImageNet-pretrained models and PSSL-pretrained models.

\end{document}